# Emotion Analysis of Tweets Banning Education in Afghanistan

Mohammad Ali Hussiny and Lilja Øvrelid
University of Oslo, Language Technology Group
{mohamhu, liljao}@ifi.uio.no

June 28, 2023

## Abstract

This paper introduces the first emotion-annotated dataset for the Dari variant of Persian spoken in Afghanistan. The LetHerLearn dataset contains 7,600 tweets posted in reaction to the Taliban's ban of women's rights to education in 2022 and has been manually annotated according to Ekman's emotion categories. We here detail the data collection and annotation process, present relevant dataset statistics as well as initial experiments on the resulting dataset, benchmarking a number of different neural architectures for the task of Dari emotion classification.

## 1 Introduction

Expression and recognition of feelings are crucial aspects of human communication and social interaction (Dolan, 2002). They significantly influence our experiences and shape our cognitive abilities, making emotional intelligence an essential component of artificial intelligence (Dolan, 2002). Emotion analysis is a growing research area that aims to enable machines to effectively recognize, analyze and understand human feelings and thinking (Mirzaee et al., 2022). Unlike sentiment analysis, emotion detection usually covers a broader range of responses, detecting a variety of emotions such as Anger, Sadness, Fear, Disgust, Happiness and more.

Online social media platforms allow people to express their views on a wide range of topics such as personal, social, political, or even commercial views. Twitter is one of the rich online sources for text analysis tasks as it is concise yet abundant in emotional context. On Twitter, communication is unrestricted by politics, age, culture, gender, and other barriers (Ghosh et al., 2020). In the current media landscape, knowledge about people's opinions and emotions as expressed on social media can be important for various objectives, such as customer service, online sale, the analysis of political and cultural events etc.

On December 20, 2022, the Taliban regime banned girls and women from pursuing education and employment in Afghanistan. This announcement shocked the world and the people of Afghanistan, and it was met with a serious and swift reaction from politicians and citizens of different countries, as well as the United Nations, political and civil figures, women activists, and citizens of Afghanistan. Many expressed their feelings against the Taliban's decision on Twitter, Facebook and other social media. In this paper, we present LetHerLearn: a Persian Dari corpus of emotion-annotated Twitter data based on the collection and analysis of tweets related to the ban of education in Afghanistan by the Taliban regime. The goal of this work is to provide insights into people's real-time perspectives, attitudes, concerns and reactions in the face of this oppression.

The paper is structured as follows. Section 2 discusses related work, focusing in particular on previous work for Persian, Section 3 then goes on to describe the creation of the LetHerLearn dataset, detailing the motivation for this work, data collection, annotation and relevant statistics. Section 4 presents details on modeling and results for experimental evaluations of a number of neural architectures trained and evaluated on LetHerLearn, and finally, Section 5 concludes the paper and describes some possible avenues for future work.

## 2 Related work

In recent years, research on emotion recognition from text has received increasing attention in the research community, and several annotated corpora have been created for this purpose (Ghosh et al., 2020). These corpora serve as valuable resources for researchers to develop and build emotion recognition models (Nandwani and Verma, 2021). While there has been significant progress in emotion recognition research from text, there are still some languages for which there is relatively little research. Persian is one such language, where there is currently not much research and limited availability of these types of datasets. Despite the relatively limited previous work on emotion detection in Persian language, there is some work on resource creation in the related area of Sentiment Analysis, such as the SentiPers dataset (Hosseini et al., 2018), the Digikala dataset (Zobeidi et al., 2019) and the Pars-ABSA dataset (Ataei et al., 2019) , all based based on Iranian user comments.

When it comes to the task of Persian Emotion Detection, the ARMANEMO dataset (Mirzaee et al., 2022) contains user opinions from social media and the dataset is annotated using a mixture of manual and automatic steps, labeling 7500 comments into the 7 classes of Anger, Fear, Joy, Hatred, Sadness, Surprise and Others. The authors trained and evaluated a number of neural models (CNN, RNN, ParsBERT, XLM-Roberta-base and XLM-Roberta-large models) on the dataset and the best performing model was XML-RoBERTa-large, achieving a macro-averaged F1 score of 75.39%. The EmoPars dataset (Sabri et al., 2021), contains 30,000 emotional tweets collected from Twitter using specific emotion-related keywords and the dataset was manually annotated into the Anger, Fear, Happiness, Hatred, Sadness and Wonder classes. This constitutes the most similar existing dataset to the one presented here. In the following we will discuss the rationale behind the data creation effort presented here.

## 3 Dataset creation

Below we detail the creation of the LetHerLearn dataset, we begin by discussing the demand that has motivated the creation of this dataset (3.1), the data collection method (3.2), continuing on to explaining the labeling and annotation process (3.3) and finally we provide some relevant statistics of our data set (3.4).

### 3.1 Demand and Importance

Despite the previous research on emotion detection in Persian, as detailed in Section 2 above, there is still a lack of research and resources for different Persian varieties. The Persian language is an Indo-European language which has more than 110 million speakers worldwide and is an official language in Iran, Afghanistan and Tajikistan (Heydari, 2019). The Persian variant spoken in Iran is called Farsi, in Afghanistan it is called Dari and in Tajikistan Tajiki (Spooner, 2012). Farsi, Dari and Tajik have the same alphabet and grammar with different accents on words in each country. There are, however, clear differences in vocabulary, where Farsi tends to have more borrowings from French and Dari from English. Crucially, however, all the described datasets above are developed based on Iranian social media and speakers and none of these are based on textual data from Afghanistan and Tajikistan. The lack of an emotion annotated dataset from Dari speakers of Persian, has motivated the creation of the Dari LetHerLearn dataset described here. As mentioned earlier, the events on December 20, 2022, where The Taliban banned education and all work activities for girls and women in Afghanistan caused massive emotional reaction on social media. We decided to base the first emotion annotated Dari dataset on social media data in order to analyse the reaction and opinion of the people faced with this event.

### 3.2 Data collection

The data constituting the LetHerLearn dataset was collected using Twitter's official developer API. We use the Tweepy library and Python language to extract Persian tweets from the Twitter API. We collected tweets using several relevant Hashtags such as #LetHerLearn, #AllOrNone, #LetHerwork, #LetAfghanistanGirlLearn and #letAfghangirllearn, which were used by Twitter users in support of the education and work for the women of Afghanistan. The included tweets were all posted from December 20, 2022 up to March 10, 2023. The search was conducted from December 20, 2022 up to March 10, 2023 and using the mentioned hashtags, we collected around fifty thousand tweets. Following removal of duplicated tweets, we selected 7600 tweets for manual labeling.

### 3.3 Data annotation

Two annotators were involved in labeling the LetHerLearn corpus. Both of the annotators are Dari native speakers with good knowledge and understanding of Dari grammar. We annotated based on Ekman's (Ekman, 1992) set of fundamental emotions, which is widely used by annotators for annotation of emotions in text. The corpus includes 6 fundamental Emotions (Anger, Disgust,

| Tweet | Label |
|---|---|
| خدا لعنت کند کسانیکه را که کاشانه و ماوایم خراب کرد | Disgust |
| May God condemn those who have destroyed our home and shelter | |
| دختر یعنی ریشه دواندن در دل خاک همیشه در حال پیشرفت و توسعه میباشد کسی دختران حذف کرده نمیتوانند | Happiness |
| A girl is like a tree, she keeps growing strong, impenetrable roots deep in the ground | |
| از ترس طالبان کسی صدایش بلند کرده نمیتواند | Fear |
| No one can raise their voice due to the dread of the Taliban | |
| احساس سوختن به تماشا نمی شود آتش بگیر تا که بدانی چه می کشم | Anger |
| Watching someone on fire doesn't truly convey any feeling, however once you experiencing the torment of being on fire, you will grasp the real pain | |

Table 1 : LetHerLearn example tweets with emotion label

Fear, Happiness, Sadness and Surprise) and we used the 'Other' category for tweets that do not fall into any of the six basic Emotions. Each tweet was assigned a maximum of one emotion. In the case of tweets containing several emotions, the annotators were instructed to assign the emotion they felt was dominant. The annotators were provided with a set of annotation guidelines written in Dari. The annotators were instructed to remove tweets in languages like Pashto and Uzbek, even if they were written in the Persian script. Incomplete tweets, for example, those missing parts of the content along with hashtags or external links, should also be removed. The full set of guidelines (in Dari and English translation) are distributed along with the dataset, however we provide a brief summary of the guidelines below.

**Annotation guidelines** The guidelines provided to the annotators contain detailed descriptions of the six emotions with example words typically associated with the different emotions. For instance, the Anger class was described as comprising tweets reflecting emotions of anger, criticism, or frustration where the text may be confrontational, express strong negative feelings, or carry a tone of harsh criticism. Words symbolizing anger might include terms such as 'lying', 'spy', 'traitor', 'hypocrite', 'oppression' etc.

In addition to instructions describing each emotion class, care was taken to delimit the class of "Other" which represents tweets that do not display any particular emotion and convey a neutral tone. For instance, tweets about mundane activities or more fact-based posts would fall under this category. Annotators were further instructed to do their best to not let personal agreement or disagreement with the opinions stated in the tweets influence the labeling process and to label without any bias or directionality. Rather, they were instructed to depart from their interpretation of the speaker's emotional state and attempt to describe it as accurately as possible using one of the provided emotion labels.

Table 1 shows some examples of tweets (with English translations) from the LetHerLearn dataset to further illustrate the annotation effort.

**Inter-annotator agreement** We further assess the consistency of annotations and measure the agreement among two annotators using Cohen's Kappa (Cohen, 1960) for the double labeling of 100 tweets. The agreement attained over the 100 tweets was 0.80.

### 3.4 Dataset statistics

The total number of words in our dataset after removal of the tweet's Hashtag, URL and Mention is 88,875 words, where 16,276 words are unique and the average length of the tweets is 4.82 words long. Figure 1 shows the occurrences of tweets for each emotion class. Examining the content of the LetHerLearn dataset, we can see that Anger is the most observed emotion, followed by Happiness, and we find that Surprise is the least observed emotion, with only 503 occurrences. The dataset was further split into train-dev-test splits using a 80:10:10 split for experimentation. Table 2 shows the detailed class-wise distribution of train, validation, and test set.

Next, and in order to get some more insight into the contents of our dataset, we examine the distribution of most frequent words per class following stop word removal, as shown in table 3 which displays the top frequent words for each of the emotion classes. We observe that some words frequently occur in all classes such as 'Taliban', 'Afghanistan', 'girls', 'women', 'everyone'. There are also clear lexical indicators associated with each class, such as 'filthy' for Disgust, 'fear' for Fear and 'pain' for Sadness. We also observe lexical items describing the cause of emotion, e.g. 'explosion' and 'arrest' for Fear and 'justice' for Happiness.

| Type | Train | Dev | Test |
|---|---|---|---|
| Anger | 1366 | 174 | 187 |
| Disgust | 462 | 50 | 57 |
| Fear | 483 | 64 | 59 |
| Happiness | 1266 | 179 | 152 |
| Sadness | 1032 | 120 | 128 |
| Surprise | 394 | 46 | 50 |
| Other | 1082 | 128 | 128 |
| Total | 6085 | 761 | 761 |

Table 2 : Data distribution for experiments

| Class | Words |
|---|---|
| Anger | 'work´, 'should´, 'islam´, 'society´ |
| Disgust | 'curse', 'tribe', 'damnation', 'filthy' |
| Fear | 'fear', 'explosion', 'escape', 'arrest' |
| Happiness | 'justice', 'hope', 'fight', 'rights' |
| Sadness | 'pain', 'close', 'forgot', 'tired' |
| Surprise | 'again', 'wish', "someday', 'men' |
| Other | 'life','world' 'further', 'iran' |

Table 3 : Frequent words in each class following removal of stopwords.

| Model | Precision | Recall | F1 |
|---|---|---|---|
| LSTM | 0.67 | 0.63 | 0.65 |
| BiLSTM | 0.66 | 0.63 | 0.64 |
| GRU | 0.65 | 0.62 | 0.60 |
| CNN | 0.66 | 0.60 | 0.62 |
| Ensemble | 0.69 | 0.64 | 0.66 |
| ParsBERT | 0.65 | 0.65 | 0.65 |
| XML-RoBERTa | **0.70** | **0.70** | **0.70** |

Table 4 : Macro Average Precision, Recall and F1 result of all models on the LetHerLearn test set.

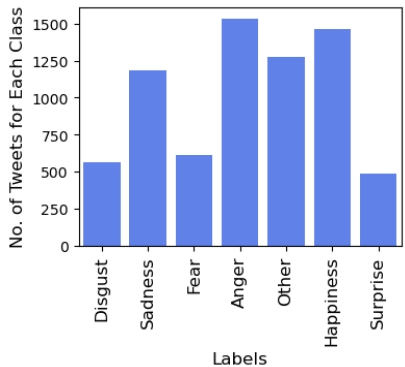

Figure 1 : Number of tweets for each emotion class in LetHerLearn.

## 4 Modeling

We evaluate a number of classic neural models on our dataset:

- Long Short-Term Memory Network (LSTM)
- Bi-directional Long Short-Term Memory Network(Bi-LSTM)
- Gated Recurrent Unit (GRU)
- Convolutional Neural Network (CNN)

All models made use of fastText (Grave et al., 2018) word embeddings with 300 dimensions for Persian. Further hyperparameters of the models are specified in Appendix A.

**Ensemble Model** After generating predicted probabilities from the LSTM, BiLSTM and GRU models, we develop an ensemble model (Dashtipour et al., 2021) using the scikit-learn library's VotingClassifier (Leon et al., 2017) class to combine the predictions result of the LSTM, BiLSTM, and GRU models.

**ParsBERT** We use a pre-trained language model for Persian, ParsBERT (Farahani et al., 2021) which is a monolingual BERT model. Hyperparameters are found in Appendix A.

**XLM-RoBERTa-large** XLM-RoBERTa is a multilingual transformer-based language model pre-trained data from over 100 different languages (Conneau et al., 2019). Hyperparameters are specified in the appendix.

### 4.1 Results

The results of our experiments are summarized in Table 4, which shows the evaluation result of the different models described above. The results show that the ensemble model achieves better results compared to the LSTM, BiLSTM, GRU and CNN models on their own, as has been shown also in previous work (Onishi and Natsume, 2014).

We further find that the XLM-RoBERTa-large model outperforms the other models. The per-class results, shown in Table 6 show that the scores vary for the different emotion classes, with the highest results obtained for the Disgust and Fear classes, and the most difficult classes being the Other class, as well as the Anger class.

| Tweet | True Label | Predicted Label |
|---|---|---|
| سنگدل سیری گرسنه های را نصیحت میکند درد گرسنگی تحمل کند<br>A hard heart satiety advise the hungry to endure the pain of hunger | Anger | Sadness |
| یکی یکی ارزو های ما نیست میشوند<br>I dreamt that my homeland had become prosperous and independent | Sadness | Anger |
| طالبان زیر فشار خارجی ها شریعت را فراموش میکنند<br>The Taliban forget the Shariah under foreign pressure | Fear | Anger |
| وای چی دردهای جانسوزی<br>Oh, What tragic and painful situation | Surprise | Happiness |

Table 5 : Examples of misclassified tweet.

| Class | Precision | Recall | F1_Score |
|---|---|---|---|
| Anger | 0.52 | 0.57 | 0.54 |
| Disgust | **0.86** | 0.84 | **0.85** |
| Fear | 0.84 | **0.86** | **0.85** |
| Happiness | 0.67 | 0.71 | 0.69 |
| Sadness | 0.58 | 0.61 | 0.59 |
| Surprise | 0.82 | 0.85 | 0.84 |
| Other | 0.62 | 0.44 | 0.52 |
| Macro Average | 0.70 | 0.70 | 0.70 |

Table 6 : Individual class performance using XLM-RoBERTa-large model.

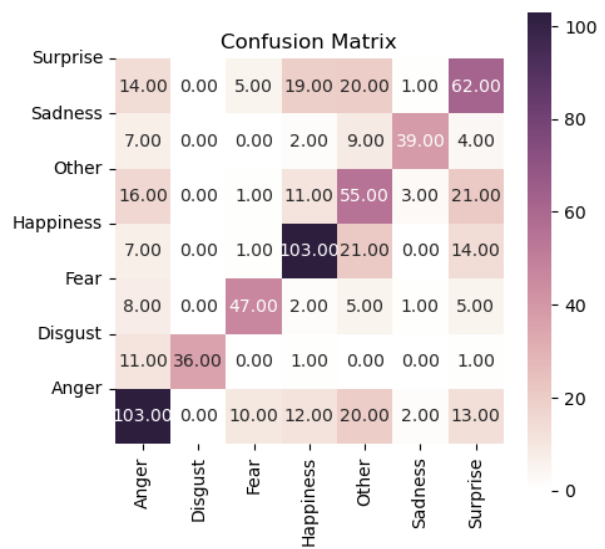

Figure 2 : Confusion matrix heat map.

### 4.2 Error analysis

We perform an error analysis on the outputs of our model in order to gain further insight into the classifications on the LetHerLearn dataset. It is clear that there is not a direct correlation between low-frequency classes (such as Disgust) and prediction performance. Figure 2 provides a confusion matrix heat map of the predictions. We find that Surprise is often mistaken for other categories, such as Happiness, Other and Anger. Not surprisingly perhaps, the Other class is also often mistaken for other classes.

Following our analysis of the misclassified predictions, we can infer some of the reasons: the assignment of a maximum of one emotion for each tweet is problematic for some of the tweets that have more than emotion. We also analyze the word overlap between the tweets and find that classes with a high degree of overlap tend to also suffer from misclassification. Table 5 shows some examples of misclassified predictions.

## 5 Conclusion

We have presented LetHerLearn: the first Dari emotion-annotated dataset of tweets collected following the Taliban's ban of women's education in 2022. All data and code will be made available.[1] In future work, we would like to experiment with cross-variant Persian emotion detection as well as multitask learning of sentiment and emotion.

---
[1] https://github.com/ltgoslo/LetHerLearn; due to the privacy restrictions applied by Twitter API, only tweet IDs along with annotations will be made available.

# References

Taha Shangipour Ataei, Kamyar Darvishi, Soroush Javdan, Behrouz Minaei-Bidgoli, and Sauleh Eetemadi. 2019. Pars-absa: an aspect-based sentiment analysis dataset for persian. *arXiv preprint arXiv:1908.01815*.

Jacob Cohen. 1960. A coefficient of agreement for nominal scales. *Educational and psychological measurement*, 20(1):37–46.

Alexis Conneau, Kartikay Khandelwal, Naman Goyal, Vishrav Chaudhary, Guillaume Wenzek, Francisco Guzmán, Edouard Grave, Myle Ott, Luke Zettlemoyer, and Veselin Stoyanov. 2019. Unsupervised cross-lingual representation learning at scale. *arXiv preprint arXiv:1911.02116*.

Kia Dashtipour, Cosimo Ieracitano, Francesco Carlo Morabito, Ali Raza, and Amir Hussain. 2021. An ensemble based classification approach for persian sentiment analysis. *Progresses in Artificial Intelligence and Neural Systems*, pages 207–215.

Raymond J Dolan. 2002. Emotion, cognition, and behavior. *science*, 298(5596):1191–1194.

Paul Ekman. 1992. Facial expressions of emotion: an old controversy and new findings. *Philosophical Transactions of the Royal Society of London. Series B: Biological Sciences*, 335(1273):63–69.

Mehrdad Farahani, Mohammad Gharachorloo, Marzieh Farahani, and Mohammad Manthouri. 2021. Parsbert: Transformer-based model for persian language understanding. *Neural Processing Letters*, 53:3831–3847.

Soumitra Ghosh, Asif Ekbal, Pushpak Bhattacharyya, Sriparna Saha, Vipin Tyagi, Alka Kumar, Shikha Srivastava, and Nitish Kumar. 2020. Annotated corpus of tweets in english from various domains for emotion detection. In *Proceedings of the 17th International Conference on Natural Language Processing (ICON)*, pages 460–469.

Edouard Grave, Piotr Bojanowski, Prakhar Gupta, Armand Joulin, and Tomas Mikolov. 2018. Learning word vectors for 157 languages. *arXiv preprint arXiv:1802.06893*.

Mohammad Heydari. 2019. Sentiment analysis challenges in persian language. *arXiv preprint arXiv:1907.04407*.

Pedram Hosseini, Ali Ahmadian Ramaki, Hassan Maleki, Mansoureh Anvari, and Seyed Abolghasem Mirroshandel. 2018. Sentipers: a sentiment analysis corpus for persian. *arXiv preprint arXiv:1801.07737*.

Florin Leon, Sabina-Adriana Floria, and Costin Bădică. 2017. Evaluating the effect of voting methods on ensemble-based classification. In *2017 IEEE international conference on INnovations in intelligent SysTems and applications (INISTA)*, pages 1–6. IEEE.

Hossein Mirzaee, Javad Peymanfard, Hamid Habibzadeh Moshtaghin, and Hossein Zeinali. 2022. Armanemo: A persian dataset for text-based emotion detection. *arXiv preprint arXiv:2207.11808*.

Pansy Nandwani and Rupali Verma. 2021. A review on sentiment analysis and emotion detection from text. *Social Network Analysis and Mining*, 11(1):81.

Akinari Onishi and Kiyohisa Natsume. 2014. Overlapped partitioning for ensemble classifiers of p300-based brain-computer interfaces. *PloS one*, 9(4):e93045.

Nazanin Sabri, Reyhane Akhavan, and Behnam Bahrak. 2021. Emopars: A collection of 30k emotion-annotated persian social media texts. In *Proceedings of the Student Research Workshop Associated with RANLP 2021*, pages 167–173.

Brian Spooner. 2012. 4. persian, farsi, dari, tajiki: Language names and language policies. In *Language Policy and Language Conflict in Afghanistan and Its Neighbors*, pages 89–117. Brill.

Shima Zobeidi, Marjan Naderan, and Seyyed Enayatallah Alavi. 2019. Opinion mining in persian language using a hybrid feature extraction approach based on convolutional neural network. *Multimedia Tools and Applications*, 78:32357–32378.

## A Appendix

### A.1 Model hyperparameters

**Long Short-Term Memory Network(LSTM)** Our model has 128 neurons with dropout and recurrent-dropout of 70%. The optimizer is adam and the number of epochs is 30 with learning rates of 0.01. and an output layer with 7 neurons, one for each class, batch size is 80.

**Bidirectional Long Short Term Memory Network (Bi-LSTM)** We use Bidirectional LSTMs with SpatialDropout1D of 0.2 and dropout and recurrent-dropout of 70%. We trained with different optimizers and achieved the highest result with adam, 30 epochs and a learning rates of 0.01 with batch size 64.

**Gated Recurrent Unit (GRU)** The GRU network has 4 layer in our proposed model and after feature extraction using fastText word embedding, the embedding layer of size (15130, 300) constitutes the input. We used adam optimizer with $\beta 1$ = 0.9, $\beta 2$ = 0.999 and learning rate of 0.01. Our model includes 64 neurons, SpatialDropout1D of 0.2 and dropout and recurrent-dropout of 65%.

**Convolutional Neural Network (CNN)** Our proposed CNN model has four layers, the Conv1D layer has 256 filters and a kernel size of 5 with relu activation function. The Dense layer has seven unit with softmax activation function, batch size is 96, Dropout value is 0.7 and we used the adam optimizer with $\beta 1$ = 0.9, $\beta 2$ = 0.999 and learning rate of 0.01.

**ParsBERT** The hyperparameters for the ParsBERT model fine-tuning was performed for seven epochs with a batch size of 32, specifying the maximum length of the encoded sequence to 128 and using the AdamW optimizer with $\beta 1$ = 0.9, $\beta 2$ = 0.999, learning rate scheduler is (2e-5) and linearly decreasing from the initial to 0 by the end of the last epoch.

**XLM-RoBERTa-large** The hyperparameters for the the XLM-RoBERTa-large is eight epochs, batch size of 32, learning_rate of 2e-5, optimizer of Adam and maximum length of 128. Table 6 shows the prediction results for each emotion class using this model.